# Skill-Based Differences in Spatio-Temporal Team Behavior in *Defence of The Ancients 2 (DotA 2)*

Anders Drachen, Matthew Yancey, John Maguire, Derrek Chu, Iris Yuhui Wang,
Tobias Mahlmann, Matthias Schubert, and Diego Klabjan

*Abstract*— Multiplayer Online Battle Arena (MOBA) games are among the most played digital games in the world. In these games, teams of players fight against each other in arena environments, and the gameplay is focused on tactical combat. Mastering MOBAs requires extensive practice, as is exemplified in the popular MOBA Defence of the Ancients 2 (DotA 2). In this paper, we present three data-driven measures of spatio-temporal behavior in DotA 2: 1) Zone changes; 2) Distribution of team members and: 3) Time series clustering via a fuzzy approach. We present a method for obtaining accurate positional data from DotA 2. We investigate how behavior varies across these measures as a function of the skill level of teams, using four tiers from novice to professional players. Results indicate that spatio-temporal behavior of MOBA teams is related to team skill, with professional teams having smaller within-team distances and conducting more zone changes than amateur teams. The temporal distribution of the within-team distances of professional and high-skilled teams also generally follows patterns distinct from lower skill ranks.

## I. Introduction and Motivation

In recent years the e-sports environment around online digital games have gained immense momentum. SuperData [1] reported a worldwide audience of 71 million people who watch competitive gaming, with 31.4 million participation or viewership in the United States. On the company side, considerable resources are being allocated to support the e-sports environment from the main companies in the domain such as Riot Games, Wargaming, Valve, Ubisoft and Turbine. In 2013, prize money for the top three tournaments (*Defense of the Ancients 2 (DotA 2)*, *League of Legends* (LoL) and *Call of Duty* (CoD) Championships) rose above 1 million USD. For DotA 2, the main tournament of the year, *The International,* contained a 10.9 million USD prize pool at the time of writing [2]. This is a tenfold increase in just 2 years and the largest in e-sports history. The prize increase was driven by an intiative by Valve, where players contributed to the prize pool by buying an in-game item *The Compendium*. In return, the compendium gave players additional ways to interact with the tournament (e.g. voting on an all-stars match) and further cosmetic items for their heroes. The *League of Legends* championship alone had estimated 32 million viewers according to [1] which is more than the finals of the National Basketball Association (NBA) in the USA. The number of players active in the gaming community is also increasing. For example, *League of Legends* alone had 52 million monthly active users in 2013 [1]. In short, MOBAs are today among the most played games in the world [3].

The e-sports environment is being driven primarily by Real-Time Strategy (RTS) games, with *StarCraft* forming the classic example, and First/Third Person Shooters (FPS, TPS) such as CoD. In recent years, Multi-Player Online Battle Arena games (MOBAs), a variant of the RTS genre have evolved to comprise a significant fraction of the e-sports environment. The term MOBA can be used in broad sense to any game where two teams consisting of several players compete with each other to achieve a victory. This is generally achieved by eliminating the opposing teams base or via a ticker system which responds to zonal or strategic point control [4,5,6]. This is a broad definition that can include popular online team-based games such as *World of Tanks* and *Call of Duty*. However, in most cases the term MOBA is used to denote a specific form of game where two teams of five players each controlling one avatar compete. The avatar is often called a hero, and gameplay is viewed from an isometric perspective. The games are played in real-time, and as real-word team-based sports, gameplay is highly dynamic.

MOBAs are complex games and thus, a MOBA features dozens or hundreds of potential heroes. Each hero has an individual set of abilities which can be developed during each match in a variety of directions by gaining experience or gold. In order to win, each team must coordinate its actions and react to the actions of the opposing team as efficiently as possible. The game is played and coordinated in a spatio-temporal way. Furthermore,,the complexity of gameplay is increased by the presence of computer-controlled objects and entities. For example, these entities may attack player-controlled heroes or provide additional experience and resources when killed. In addition, MOBAs feature a variety of items that can provide different bonuses or penalties to heroes. In essence, MOBAs feature a broad spectrum of possible behaviors which means that mastering these games is quite challenging. The combat emphasis means that players and teams must have a strong tactical knowledge and invest substantial time to master the game, similar to comparable off-line sports. In summary,

MOBAs are of interest from the perspective of sports analysis [5] and game analytics [7] in general as a major e-sport, backed by a growing industry. Given the relative dearth of analytical work on e-sports, with notably exceptions e.g. from the domain of game AI [8,9,10], MOBAs remain relatively unexplored from an analytical point of view. One of the central aspects of MOBAs is team behavior, and how this behavior varies as a function of player skill [5,6]. In this paper, we present three different analyses of player behavior in the MOBA DotA 2, focusing on differences in spatio-temporal behavior as a function of team skill. The work presented builds towards lines of tactical analysis permitting the analysis of team strategies in the game.

## II. CONTRIBUTION

In this paper, we investigate how spatio-temporal behavior of teams in DotA 2 varies across four skill tiers and across two spatial measures: zone changes (changes in position in the game as a function of the underlying terrain) and team distribution (the distance between heroes on a team). The main contributions are as follows:

*1)* A method for extracting precise spatio-temporal information from DotA 2 replay files, enabling researchers to perform detailed analysis of team behavior. The parser used (Dotalys 2) also provides associated event data, and allows for visualizing player trajectories.

*2)* Three different analyses of spatio-temporal team behavior, focusing on zone changes and team distribution, across four skill tiers. The results show that there are significant differences in the behavior of teams across skill tiers for the investigated features.

Replay files (game logs) from 196 matches across four skill tiers were used as the basis for this analysis. Results indicate that spatio-temporal behavior of MOBA teams is related to team skill with professional teams having smaller within-team distances and conducting more zone changes than amateur teams. The temporal distribution of the within-team distances of professional and high-skilled teams also generally follows patterns distinct from lower skill ranks. The fundamental goal of the work is similar to that of Yang et al. [6] and focuses on the behavior in MOBAs. In addition, a further goal is to aid DotA 2 players in visualizing, analyzing and improving their performance by exploring spatio-temporal behavioral data.

## III. RELATED WORK

In recent years, there has been a wealth of publications that analyse player behavior in games for a variety of reasons and within several domains, including game AI [9-11], game development [7], networking [12], e-sports [5], business [13], social computing [14], learning/serious games, for the analysis of player behavior directly [15-17] and more. A full review of these domains is out of scope of this paper. Focus will here be on work directly on MOBAs or related to the analyses presented.

MOBAs have a varied history. One of the first games in this Genre was the original *Defense of the Ancients* (DotA) which is a player-developed mod for *Warcraft 3: Reign of Chaos*. Warcraft 3 is an RTS which provided a powerful map editor and added the concept of hero units into the RTS genre. Thus, it allows to design maps mimicking the style of player vs. player (PvP) instanced battlegrounds being common in Massively Multi-Player Online Games (MMOG) such as *World of Warcraft* and *Guild Wars*. Analysis work on MOBAs is relatively rare and first approaches where only published very recently. Thus, even the recent review on spatio-temporal game analytics by Drachen and Schubert [18] does not include the newest work in the area. However, some of the previous research in digital games is directly relevant to the contemporary research and bears mentioning. Miller and Crowcroft [4] examined character movement in the *Arathi Basin* PvP battleground of the MMOG *World of Warcraft*, obtaining data via mining the server-client stream (a method also used by e.g. Shim et al. [19]). The *Arathi Basin* is a 10v10 arena-style instanced challenge that pits two teams against each other trying to control five strategic points to reach a certain amount of resources before the other team does. The authors examine three types of models: 1) Waypoint based movement models, 2) Spatial hot spots, and 3) Flocking behavior of the avatars. The authors concluded that there were distinct patterns in the spatial behavior of the individual players, labelling several of these according to the in-game purpose of the behavior. E.g. guardians, who stayed close to a base (strategic point) for the duration of a match to defend it. The authors also examined team-based movement. The authors recorded characters moving together within a 30 yards diameter (in-game years) for several seconds. However, this analysis did not reveal typical group movement, which was explained by dying players which respawn at the nearest graveyard (the point where dead players rejoin the game) and rejoin different groups of other players. This result is partly contrasted by Rioult et al. [5], who worked with DotA (1) replay files. Replay files are generated by online game clients for the purpose of sharing a match with others. Replay files basically allow another players client to replay the match. Thus, replay files contain highly detailed information about matches. Thanks to the player community sharing these files online, for example via tournament sites or sites such as gosugamers.com, thousands of replay files are available for download. This forms the source of information for current work on MOBAs. Rioult et al. [5] used topolocial measures - areas of polygons described by the players, inertia, diameter, distance to the base – to show that outcomes of matches can be predicted with recall rates of 90%. The authors concluded that topological clues appear relevant for predicting outcomes of MOBA matches as well as for determining which features that are relevant to drive prediction, e.g. team aggressiveness. A potential limitation of the work presented is that spatial positioning information was estimated via movement orders triggered by players, which adds an unknown amount of uncertainty to the positional data used. The authors highlight that mining player trajectories in e-sports is of interest in offline sports as well. Schubert et al. [20] further noted that the high spatio-temporal precision of measures in digital games is often superior to what can be achieved in comparable real-life settings.

Batsford [21] used a feed-forward sigmoidal neural network in connection with several genetic algorithms to discover optimal routes in the jungle areas of DotA 2 (see below), using simulation. The author noted that there appeared to be convergence towards optimal jungle routes, but that more work was needed and highlights the complexity involved in representing the game state in DotA 2. Yang et al. [6] worked with replay files from professional players of DotA 2 and presented an approach for discovering and defining patterns in combat tactics including the temporal but not spatial information. The authors defined specific roles of DotA 2 players in the game, modelled combat as a sequence of graphs and used this representation to extract patterns that predict successful outcomes of individual fights in the game as well as matches with an 80% prediction accuracy. The authors attributed features to the graphs using frequent sub-graph mining which allowed them to describe how different combat tactics contributed to team success in specific situation.

On the topic of MOBAs, but not focused on in-game behavioral analysis, Nuangjumnonga and Mitomo [22] analyzed results from a close-ended survey to examine potential correlations between behavior and leadership development in the MOBAs DotA 2 and Heroes of Newerth. The authors noted that different roles in the game require different kinds of leadership. Pobiedina et al. [23] used data from the DotA 2 web community to investigate teamwork, concluding for example that more experienced teams win more often, and that playing with in-game friends as opposed to pick-up teams increases the chance of winning. Harrop [24] investigated the nature of rules in DotA, and noted that players use difference "truce calls" to negotiate rules and the maintenance of fair play in the game. These works operate outside the perspective of in-game spatio-temporal analysis presented here and by related work such as Yang et al. [6] and Rioult et al. [5], that mine replay files to analyze behavior during matches, whether spatio-temporal or not.

Outside the MOBA domain, a significant amount of work has been done on spatio-temporal behavior analysis in games. A recent review is provided by Drachen & Schubert [18]. In relation to the work presented here, trajectories in 3D games have been used to direct AI bot movement in games, covering both unit movement in RTS games and bot movement in FPS games [e.g. 25]. Other works have focused on analyzing and visualizing spatial and/or spatio-temporal player behavior for the purpose of informing game design or –development [18]. For example, Hoobler et al. [15] developed a tool for visualizing player movement in team-based FPS games. Drachen & Canossa [16] presented several techniques for analyzing and visualizing competitive behavior in the same type of games. Dankoff [26] presented a description of how Ubisoft used trajectory information during user testing to evaluate the design of *Assassins´ Creed 3*. Player profiling is a key topic in both commercial game development and game AI, as noted by Bauckhage et al. [17] who investigated several spatio-temporal clustering algorithms for the purpose of profiling spatio-temporal player behavior. Finally, matchmaking based on player skills, network latency and other issues, remains one of the key challenges in competitive digital games [27].

## IV. DEFENCE OF THE ANCIENTS 2

Defense of the Ancients 2 (DotA 2) is a Free-to-Play (F2P) game that originate in the DotA mod of Warcraft 3: Reign of Chaos which was released in 2003, and was designed by an anonymous modder known as "Eul". The mod became popular and supported a substantial e-sports scene. By 2009, Valve began developing DotA 2 as a standalone game. It went into beta testing in 2011. DotA 2 is today the most played game on the online distribution platform Steam, in itself on of the most heavily trafficked digital game platforms in the world. According to SteamPowered (http://store.steampowered. com/stats/), the native analytics service of Valve, DotA 2 is played more than 1140 years every day. This is seconded by Sifa et al. [28] and Orland [29], the latter who estimated a total playtime for DotA 2 of 430,000 years by April 2014, less than a year after launch in July 2013, but also after a prolonged beta period lasting since 2011 (Fig. 1). Dota 2 also hosts the biggest prize pools in e-sports. Furthermore, the game has a reported number of 7.86 million of monthly active players and achieved a revenue of 80 million USD in 2013 via micro-transactions which are limited to selling cosmetic items. These numbers should be compared to League of Legends, the most popular game in the MOBA genre, with an estimated 67 million monthly active users in 2013 and 624 million USD revenue in the same year [30].

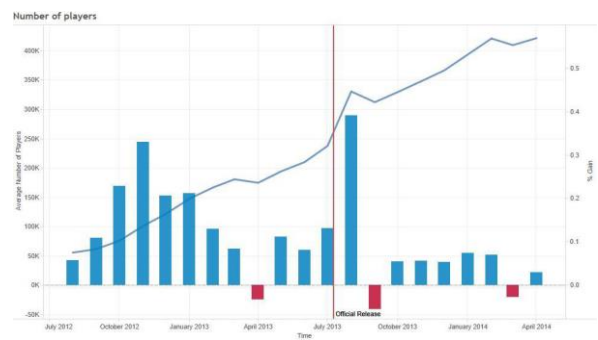

Fig. 1: The number of DotA 2 players since July 2012 (blue line). Gain in % is shown in the bars along the X-axis (data courtesy of Valve via http://steamcharts.com/app/570#All).

**Gameplay:** DotA 2 is played by two teams of 5 players, each of them controlling one avatar-character being selected from a roster of 112 (at the time of writing) predefined heroes. Each hero has different abilities and is suited for different types of roles in the game, e.g. for dealing damage at a distance. Each hero can gain levels similar to an archetypical character in a Role-Playing Game (RPG). Additionally, a hero can be equipped with a variety of objects that improve the characters base statistics or increase, alter or add new abilities. Items are bought with gold being earned during the game, e.g. from killing other player's characters, creeps or towers (see below). Experience to level up characters and thereby unlock better versions of the hero´s abilities or new abilities, is earned in a similar way. Tactics and strategy are key components in the game, and communication between team members is very important. Players can communicate via text chat, voice chat, alert messages in the arena itself ("pings") or by writing on the minimap, which provides an overview on the games UI. The game is viewed from an isometric perspective. Games have no time limit, but the matches used in the current work average about 40 minutes in length. The two teams compete in a geographically balanced, square virtual arena, and the same arena is used in every match. The arena is split in two parts, with each half owned by one team at the beginning of a match (Fig. 2). The arena contains a variety of game-related features, most importantly a base for each team with a central building, the ancient, which the opposing team must destroy to win. The ancients are guarded by a series of defensive structures called towers which provide offensive or defensive capabilities. Additionally, the two bases regularly spawn computer-controlled units called creeps which rush the opposing team's towers and players. The presence of towers and creeps results in an unstable balance that oscillates slowly (Rioult et al., 2014). There are three main pathways through the map, referred to as lanes (Fig. 2). These are differentiated "top", "middle", and "bottom". The lanes form vital strategic points of attack on the opposing team's defenses. However, there are a variety of sub-environments in the DotA 2 environment, which sees different tactical and strategic uses. For example, the jungle area in between the lanes form a means for leveling up a hero via killing regularly re-spawning computer-controlled neutral units, as well as for launching surprise attacks on enemy players or creeps.

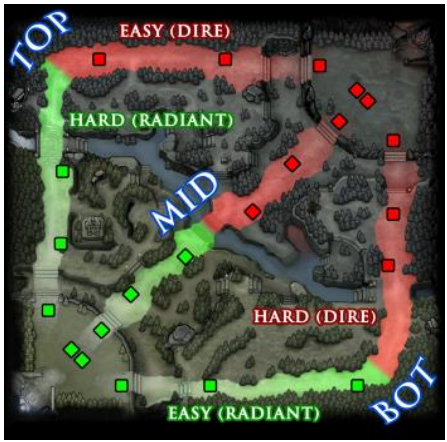

Fig. 2: The DotA 2 map. The Radiant base is at the bottom left, the Dire base at the top right. The three lanes are indicated as is the location of towers (see text) (Source: the DotA 2 wiki, http://dota2.gamepedia.com/Lane).

V. DATA AND PRE-PROCESSING

Thanks to the ability of the DotA 2 game to generate replay files and the DotA 2 community sharing these files, replays are publicly available. Furthermore, the e-sports community adds replays from tournaments. The files are available via online libraries such as www.gosugamers.com.

DotA 2 replay files are collected by Valve, and can be accessed directly through the game client, which uses a native interface to communicate with the DotA 2 master server. Replay files are used by DotA 2 players/teams for tactical analysis of matches, as well as an analysis tool of the overall game mechanics. For example, sites like www.dotabuff.com collect information about the popularity and the winning percentages of each hero using a web API.

In this paper, match IDs and associated Matchmaking Ratings (MMRs, see below) were collected manually from 50 matches across four different skill tiers. The corresponding replays files were then downloaded. Data from 200 matches in total were collected in total, however, for 4 of these players on either the Radiant or Dire team left the game very early, unbalancing the gameplay, rather than finishing the match. These games were removed from the dataset. The time ticker in the replay files were standardized to the nearest second. As each player can have different time stamps for the period of a match, standardization is required in order to conduct across-match analysis. It is important to note that the data collected are from matches from the Spring 2014. DotA 2 is continually updated, and it is therefore possible that patterns described here will change in the future. For example, strategies among professional teams change regularly, as evidenced in the primary DotA 2 tournament *The International* from the year 2013 to 2014.

The division of matches into skill brackets was based on corresponding tiers of MMRs. MMR is a metric that Valve uses to sort players of the same skill level, for matchmaking purposes. The approach is similar to ELO method for calculating the relative skill levels of players in competitor-vs-competitor games such as chess. The following four brackets were considered: MMR 2,000-3,000: Normal tier; MMR 3,000-4,000: High Skill tier; MMR > 4,000: Very High Skill tier. The Professional tier was given to players from professional tournament matches. There are no MMR for these matches, but are only associated with professional, fulltime players whose level of skill surpasses those in the 4,000+ MMR bracket.

Several community-developed parsers were investigated for extracting event-based data, notably Bruno´s [31]. This parser is however not compatible with recent replay files. However, none of these can provide spatio-temporal information. Therefore, the primary parser used was Dotalys developed by Tobias Mahlman [32]. Dotalys was further developed for the purposes of the current project into Dotalys 2, adding several functionalities of which the key one is the ability to extract precise spatio-temporal information from DotA 2 replay files, and thereby player trajectories (Fig. 3). Additionally, several options for visualizing the information in replay files, or groups of replay files, were added, including dynamic trajectory heatmaps. Two key problems needed to be solved before spatio-temporal analyses of player behavior can be performed in DotA 2: 1) spatio-temporal information about hero positions and stationary objects in the game need to be obtained; 2) the DotA 2 map need to be encoded according to the functionality of the underlying terrain.

**Spatial coordinates:** Given that other work on MOBAs, e.g. Rioult et al. (2014) uses inferred or no positional information, and the direct use of being able to map player behavior spatio-temporally in future MOBA work, the procedure of acquiring spatial coordinates is described in some detail here. A DotA 2 replay file represents a recording of a match. The game happens in ticks, with each tick representing 33 ms of wall time. The game engine models the game environment in an object-oriented manner: each game object is referred to as an entity and follows a class-based structure. Most objects on the DotA 2 map are derived from the feature *DT_BaseEntity*, found in the replay files. Heroes, for example, are referred to as *CDOTA_Unit_Hero_\** (\* being the hero name). During a state update, the DotA 2 engine stores the changes of entity properties. This information is hidden in the replay files, in that a tick packet (termed a frame), can contain multiple messages. For example user messages such as chats and disconnects: game events such as hero deaths, and similar messages which are easily parsable due to being in text format. This is the information used by e.g. Bruno's parser [31]. Messages called "packed entities" however, contain changes for a frame. These are encoded in binary and optimized for bandwith usage, i.e. packed. In some cases the packs contain four different values in one float, which leads to needing to unwind four different bytes. These packet entities can however be decoded (for instructions please refer to Valve Corporation [33]. This way, it is possible to identify the entity that has changed, including its class and the name of the property, including its new value and value class. Subsequently, reverse engineering can be applied to discover what the different classes do. *DT_BaseEntity*, introduced above, provides the properties necessary for obtaining spatial information about hero position, notably: *m_cellX*, *m_cellY*, and *m_cellZ*. X and Y correspond to a grid that is layered on top of the DotA 2 map, with the origin (0,0) in the middle of the map. Z only changes when the player walks into the river section of the map, or up on a cliff. This dimension was not used in the present work.

Positions of heroes and other moveable entities in DotA 2 are represented in a two-level hierarchy. The first level consists of a 128*128 grid overlay on the map while the second level stores the relative position of a character within the grid cell. The latter is determined by a float vector component, *m_vec_Origin*. Here the grid is used, as the analyses do not require the ability to investigate micro-patterns in player positioning. Dotalys 2 is to the best knowledge of the authors the only (now publicly available) parser that permits the extraction of spatial information from DotA 2 for hero movement (some parsers allow the generation of e.g. heatmaps of death events, based on event data). For the purposes of analyses, Dotalys 2 was modified to provide output files containing the following information: *matchID*, *team* (Radiant or Dire), *playerID*, *time stamp* and *x-y coordinates* of the position of all heroes.

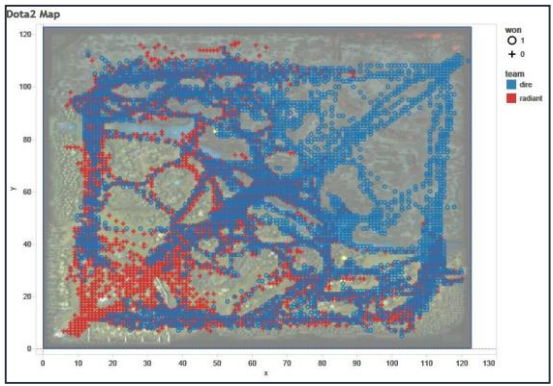

Fig. 3: Example of grid-based positional information extracted from a DotA 2 replay file, and visualized using the Dotalys 2 tool. Blue indicates positional information for the Dire team, Red indicates positional information from heroes on the Radiant team. Dotalys 2 can also export individual player trajectories

**Overlay map generation:** As outlined above, here we work with the DotA 2 grid for the purpose of analyzing and comparing movement patterns. To define certain areas in the map, we employed a two-step approach. We firstly derived grid cells being accessible by player characters and then manually labeled each accessible cell with a zone label. Secondly, to find out which grid cells can be visited by a player hero (character), we collected the grid coordinates of all ten heroes in 10 complete matches. The resulting grid was plotted into a bitmap for visual verification. After manual checking, the borders of 11 zones on the map were edited in the bitmap and the zones where color-coded. In other words, each zone has a particular color and all pixels being painted in this color belong to the zone. Now, parsing the bitmap allows us to read the color of a pixel corresponding to a particular grid cell. By reading the color of a pixel the corresponding grid cell can be mapped to a zone tag. We distinguish the following zones: *base_Radiant*, *base_Dire*, *river*, *jungle*, *lane_Shop*, *secret_Shop*, *top_Lane*, *middle_Lane*, *bottom_Lane*, *pit* and *void* (unaccessible areas) (Fig. 4).

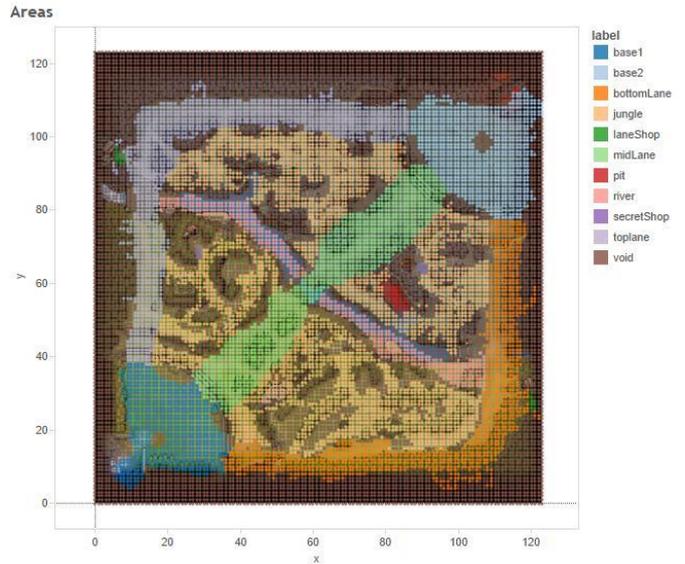

Fig. 4: grid map of the different environmental zones in DotA 2 based on 10 complete replays.

## VI. ANALYSES AND RESULTS

Three different analyses were performed on the DotA 2 dataset. Each targets a different dimension of spatio-temporal team behavior in the game across the four used skill brackets.

### A. Zone changes

The first experiment focused on investigating general movement on the DotA 2 map, as a function of the four skill tiers. The goal was to evaluate whether highly skilled players move more between the 11 navigable zones in the game. The hypothesis relates to conventional wisdom in the DotA 2 community suggesting that expert players will change zone more often than novice players in order to initiate coordinated strategies, whereas novice players will tend to stay in one of the three lanes for the major part of a match. Additionally, it is of interest to see how zone changes contributed to victories.

Using the grid map with zonal information, the trajectories of the players were assigned to a zone for each timestamp. The number of times each player changed zone was calculated, with the baseline rule that a zone change must have lasted 5 seconds to be included. This limit was imposed to avoid counting situations where players operating on the boundary of a zone might leave it for a very short interval, without doing so in gameplay terms. For example, a player operating in the middle lane, but for a second crossing the border into the jungle, before returning to the grid cells covering the lane.

A one-way ANOVA was used to test the difference in average team distance across entire matches. Intra-team distance differed significantly across the four skill tiers, F = (67.084), p < $2.2e^{-16}$ for the differences across tiers, and F = (135.802), p < $2.2e^{-16}$ for the differences across win/lose conditions. The results (Fig. 5 (top)) thereby indicate that there is a difference in the number of zone changes per minute across the four skill tiers, with the professional tier observing the most zone changes per minute. The same pattern is observed when considering winning and losing teams, across the four skill tiers (Fig. 5 (bottom)).

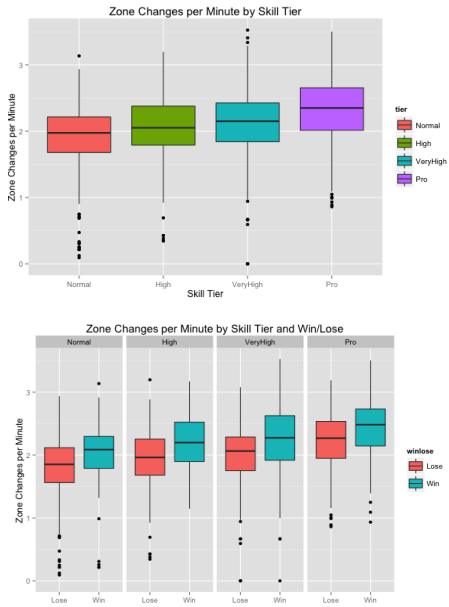

Fig. 5: (top) Box plot of the differences in the number of zone changes per minute in DotA 2 for four different skill tiers. (bottom) Box plot of the differences across the skill tiers divided according to a win/loss criterion.

### B. Team Distance

Another way to investigate team movement and teamwork behavior is to consider the distances between each member of a team. Here, the goal is to evaluate the degree of team distribution over the map and whether it is related to the skill tier and win/loss conditions. To obtain a baseline metric for intra-team distance, the average over all pairs of distances between the players on a team $T = \{p_1,...,p_n\}$ is computed. Formally, this is described in the following formula (1):

(1) $D = \frac{2}{n(n-1)} \sum_{i=1}^{n} \sum_{j=i+1}^{n} \|p_i.pos - p_j.pos\|_2$

Where $\{p_1,...,p_n\}$ are the players in team T, $n$ is the number of players on a team, and $D(T)$ is the average Euclidian (physical) distance between pairs of team member positions $p_i.pos$. Positions refer here to placement on the DotA 2 grid. This formula only requires the upper triangular part of the square distance matrix and therefore, normalization is done by dividing by *n(n-1)/2*. An alternative formulation is (2), using distance [*dist*] directly.

(2) $D = \frac{1}{n} \sum_{i}^{n} \sum_{j \neq i}^{n} dist(p_i, p_j)$

A one-way ANOVA was used to test the difference in average team distance across entire matches. Intra-team distance differed significantly across the four skill tiers, F = (62.758), p < $2.2e^{-16}$. The results indicate that there is a significant difference in the average team distance between skill tiers. The professional tier in particular is low in this metric. The team distance for professional skill level not only has a low average but also a smaller variance. This shows that professional teams are more consistent in team spread. The normal (novice) tier has the highest average and largest variance. During a DotA 2 match, there are different typical phases of the game. For example, strategies are often referred to as early or late game strategies in the DotA 2 community. In order to evaluate if team distance differences change during the three main phases of a typical match (early game, 0-15 minutes; mid game, 15-30 minutes and late game, 30 minutes and beyond), the distance data for the eight categories (4 skill tiers, win/loss) were plotted on graphs, using one second moving averages (Fig. s 6-8).

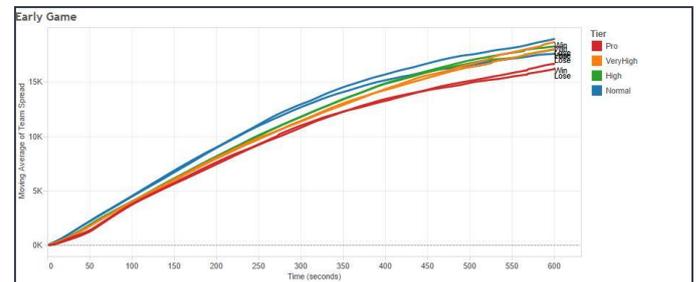

Fig. : One second moving average of the average team distance (team spread) for the early game phase of the DotA 2 matches in the dataset, plotted against match time.

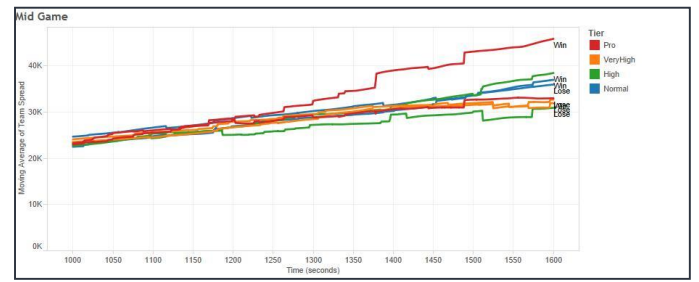

Fig. 7: One second moving average of the average team distance (team spread) for the mid game phase of the DotA 2 matches in the dataset, plotted against match time.

Results indicate that during the early game, there is a constant increase in team distance across all categories. This shows the typical opening strategies of players moving into different lanes and the jungle area. The professional tier has the smallest moving average (Fig. 6). For the mid game, the differences in intra-team distance become noticeable, with the winning-professional category showing the largest team distributions (Fig. 7). For the late game phase, the main discrepancy is observed between winning and losing teams at the professional skill tier (Fig. 8).

The explanation for this pattern may be the ability of professional teams to capitalize on errors by the opposing team at crucial moments, allowing for small margins of error in performance.

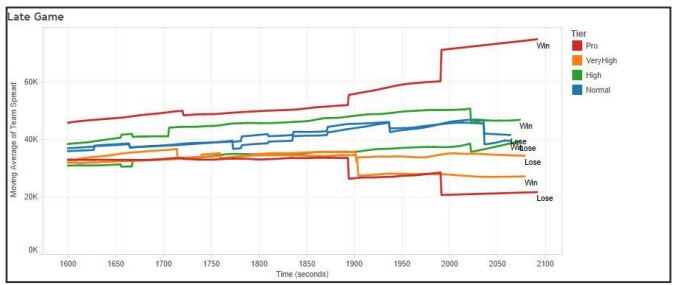

Fig. 8: One second moving average of the average team distance (team spread) for the late game phase of the DotA 2 matches in the dataset, plotted against match time.

## C. Time-Series Clustering

To identify more intricate and prevalent trends in our data, we sought to utilize unsupervised learning via time series clustering of average distance between players per second. Cluster analysis has in the past been used to find patterns in non-spatial player behavior [18,28] The objective is two-fold: to find matches where players exhibit similar movement patterns, and to explore what factors may lead to different movement behaviors.

The initial challenge is to define the concept of "similarity" between time series. Varying match lengths and high dimensionality meant that conventional distance metrics such as Manhattan and Euclidean Distances would not be appropriate due to e.g. scaling issues. Relevant literature suggested using Dynamic Time Warp (DTW), a popular technique utilized in speech recognition, as a robust similarity measure that could account for time series of different length [34]. However, conventional DTW algorithms for time series of length n and m run in O(nm), and creating a distance matrix for over 300 times series of length > 1000 would be computationally inefficient. Therefore, DTW was rejected in favor of Permutation Distribution (PD). The PD of a time series borrows concepts from combinatorics, and can loosely be defined as a measure of the complexity of a time series, where similarity is determined by the divergence between the distributions of two time series. This metric is invariant of both scale and time, and computationally efficient, running in linear time [35]. Following additional pre-processing where matches with any leaving players, even if leaving happened a few seconds before the end of the match, were removed, the dataset consisted of 48 matches for the Normal and High skill tiers, and 49 matches for the Very High tier. Finally, 45 professional tournament matches were used, for a total of 190 matches and 380 time series (two time series, Radiant and Dire teams, for each match).

We computed the Permutation Distribution of each time series with the "pdc" package in R [36] and applied k-medoids and fuzzy clustering algorithms on the resulting distance matrix with the "cluster" package [37]. Diagnostics for each set of clustering solutions were conducted by comparing resulting silhouette plots and average silhouette widths [38]. The best solution was determined with 3 fuzzy clusters with a membership exponent set at 1.15, for an average silhouette width of 0.42. This was in contrast to the best k-medoids solution, with 3 clusters and an average silhouette width of 0.34. The clustered time series were plotted against each skill tier and match outcome (win, lose). (Fig. s 8,9).

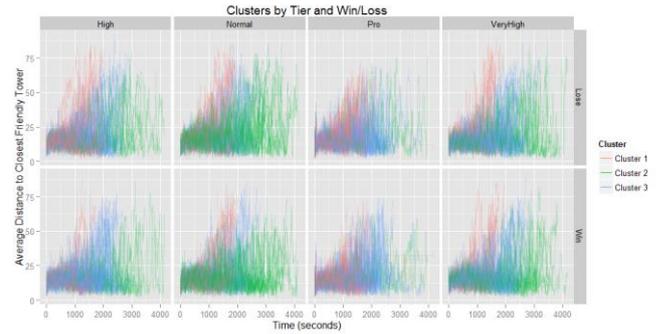

Fig. 9: Results of cluster analysis of time-series data on team distance for DotA 2 teams.

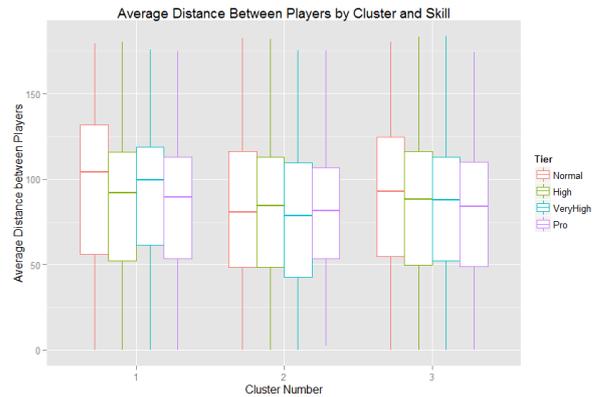

Fig. 10: Boxplot of the average distance values of each cluster, faceted by tier, demonstrating the patterns in player movement between and within each cluster.

Mean values of each time series are 87.43, 80.38, and 82.84 respectively for Clusters 1, 2, and 3, with corresponding variances of 1,507.48, 1,516.99, and 1,492.95. **Cluster 1** is comprised of matches of generally shorter length, with an average time of 1,576 seconds. **Cluster 2** is comprised of abnormally longer matches by DotA 2 standards, with an average time of 3,356 seconds, while **Cluster 3** is comprised of average length matches, with a mean time of 2,305 seconds. Sevweral prominent trends are apparent, notably that professional players exhibit lower variance of average distances to each teammate, and with the exception of Cluster 2, the absolute lowest average distances. This difference in the time series of professional and lower skill players is most pronounced in the shorter matches of Cluster 1, suggesting that "decisive" behaviors leading to early match conclusions are different across skill tiers (Fig. 9,10). However, in the longer matches of Cluster 3 and the longest matches in Cluster 2, we observed that the differences between skill brackets are less pronounced. Means and medians are more tightly coupled, with standard deviation inversely proportional to skill level in these brackets. This can perhaps be attributed to the fact that focusing on the game objectives becomes more evident to all players - regardless of skill level - as the game progresses, and team cooperation in the form of coordinated movement is increasingly encouraged or apparent as a means to victory.

## VII. CONCLUSIONS AND DISCUSSION

Multiplayer Online Battle Arena (MOBA) games are among the most played digital games in the world and support substantial e-sports communities [1,30]. In this paper, three analyses were presented focusing on spatio-temporal behavior of DotA 2 teams, across four skill tiers and two measures: **zone changes** (changes in position in terms of terrain) and **intra-team distance** (the distance between heroes on a team). The results show different ways in which the behavior of teams across skill tiers varies. Additional contributions are made in terms of a method for obtaining and visualizing accurate spatio-temporal data from DotA 2 [32]. The work presented extend previous work on MOBAs, and highlight the use of spatio-temporal patterns to analyze gameplay [4-6]. As noted in the introduction, the basic goal of the work presented here is to explore MOBAs, and work towards results and tools that will aid DotA 2 players in visualizing, analyzing and improving their performance.

The work presented is not intended as the final word on skill-based strategy analysis in MOBAs, but rather as a first step. Recent related work has shown there are a variety of ways to approach the problem of strategy description, evaluation and prediction in MOBAs [e.g. 5,6], and current work has only begun to tap into the rich and varied behavioral data available from MOBAs. Future work will include investigating match properties not tied to time series and investigate if they are also captured by the team distribution clusters. We are also investigating analyze tactical gameplay analysis using encounter detection and sequence mining. Finally, tactical behavior changes over time form a venue of future research, as evidenced by the changes in tactics used by professional teams in DotA 2 tournaments.

## REFERENCES

[1] SuperData. "esports Digital Games Brief", Digital Games Brief, 2014 [Online]. Available: http://gallery.mailchimp.com/a2b92079991 31347c9c0c44ce/files/SuperData_Research_eSports_Brief.pdf

[2] Valve Corporation. "The International DOTA2 Championships official website," 2012. [Online]. Available: http://www.dota2.com/international/overview/

[3] J. Gaudiosi. "Riot Games' League Of Legends Officially Becomes Most Played PC Game In The World," Forbes, 2012. [Online]. Available: http://onforb.es/MVqpYv

[4] J. Miller and J. Crowcroft, "Group Movement in World of Warcraft Battlegrounds," in Int. J. of Advanced Media and Communication, vol. 4, no. 4, pp. 387–404, 2010.

[5] F. Rioult, J.-P. Metivier, B. Helleu, N. Scelles and C. Durand. "Mining Tracks of Competitive Video Games," in *Proc. AASRI Conference on Sports Engineering and Computer Science,* 2014.

[6] P. Yang, B. Harrison and D. L. Roberts. "Identifying Patterns in Combat that are Predictive of Success in MOBA Games," in *Proc. Foundations of Digital Games*, 2014.

[7] Seif El-Nasr, M.; Drachen, A. and Canossa, A. (eds): *Game Analytics: Maximising the Value of Player Data*. Springer Publishers, 2013.

[8] P. Yang and D. L. Roberts. "Extracting human-readable knowledge rules in complex time-evolving environments," in *Proc. of The International Conference on Information and Knowledge Engineering*, 2013.

[9] B. G. Weber and M. Mateas. "A data mining approach to strategy prediction," in *Proc. IEEE Symposium on Computational Intelligence and Games*, pp. 140-147, 2009.

[10] Yannakakis. G. N. "Game AI Revisited," in *Proc. of ACM Computing Frontiers Conference*, 2012.

[11] M. Stanescu, S. P. Hernandez, G. Erickson, R. Greiner, and M. Buro. "Predicting army combat outcomes in starcraft," in *Proc. Ninth AAAI Conf. on Artificial Intelligence and Interactive Digital Entertainment*, 2013

[12] W.-c. Feng, D. Brandt and D. Saha. "A Long-Term Study of a Popular MMORPG," in *Proc. Netgames*, 2007.

[13] T. Fields, T. and B. Cotton. *Social Game Design: Monetization Methods and Mechanics*. Morgan Kauffman Publishers, 2011.

[14] B. Medler. "Player Dossiers: Analyzing Gameplay Data as a Reward," In *Game Studies*, 2011.

[15] N. Hoobler, G. Humphreys and M. Agrawala, M. "Visualizing competitive behaviors in multiuser virtual environments," in *Proc. IEEE Visualization Conference*, 2014.

[16] A. Drachen and A. Canossa. "Evaluating Motion: Spatial User Behavior in Virtuel Environments." In *International Journal of Arts and Technology*, vol. 4(3), pp. 294-314, 2011.

[17] C. Bauckhage, R. Sifa, A. Drachen, C. Thurau, C. F. Hadiji. "Beyond Heatmaps. Spatio-Temporal Clustering using Behavior-Based Partitioning of Game Levels," in *Proc. IEEE Comp. Intelligence in Games*, 2014.

[18] A. Drachen and M. Schubert. "Spatial Game Analytics and Visualization," in *Proc. of IEEE Computational Intelligence in Games*, 2013.

[19] K. J. Shim, N. Pathak, M. A. Ahmad, C. DeLong, Z. Borbora, A. Mahapatra, and J. Srivastava. "Analyzing Human Behavior from Multiplayer Online Game Logs: A Knowledge Discovery Approach," In *IEEE Intelligent Systems*, VOl. 26(1), 2011.

[20] Schubert, M.; Kriegel, H.-P. & Züfle, A. "Managing and Mining Multiplayer Online Games". In *Proc. SSTD*, pp. 441-444, 2011.

[21] T. E. Batsford. *Calculating Optimal Jungling Routes in DOTA2 Using Neural Networks and Genetic Algorithms*. Student paper, University of Derby (UK) [Online]. Available: http://commerce3.derby.ac.uk/ojs/index.php/gb/article/view/14

[22] T. Nuangjumnonga and H. Mitomo. "Leadership development through online gaming". In *Proc. 19th ITS Biennial Conference*, International Telecommunications Society, 2012.

[23] N. Pobiedina, J. Neidhardt, M. d. C. Calatrava Moreno, and H. Werthner. "Ranking factors of team success". In *Proc. of the 22$^{nd}$ International Conference on World Wide Web Companion*, pp. 1185-1194, 2013.

[24] M. Harrop. "Truce in Online Games," in *Proc. of OZCHI*, pp. 297-300, 2009.

[25] C. Bauckhage, M. Roth, and V. Hafner, "Where am I? – On Providing Gamebots with a Sense of Location Using Spectral Clustering of Waypoints," in *Proc. SAB Workshop on Player Satisfaction*, 2006.

[26] J. Dankoff. "Game Telemetry with Playtest DNA on Assassin's Creed," Gamasutra, 2014. [Online]. Available: http://www.gamasutra.com/blogs/JonathanDankoff/20140320/213624/Game_Telemetry_with_DNA_Tracking_on_Assassins_Creed.php

[27] O. Delalleau, E. Contal, E. Thiboudeau-Laufer, R. Ferrari, Y. Bengio and F. Zhang. "Beyond Skill Rating: Advanced Matchmaking in Ghost Recon Online". In *IEEE Transactions on Computational Intelligence and AI in Games,* vol. 4(3), pp. 167-177, 2012.

[28] R. Sifa, C. Bauckhage and A. Drachen. "The Playtime Principle: Large-scale Cross-games Interest Modeling". In *Proc. IEEE Computational Intelligence in Games*, 2014.

[29] K. Orland. "Introducing Steam Gauge: Ars reveals Steams most popular games," Ars Technica, 2014. [Online]. Available: http://arstechnica.com/gaming/2014/04/introducing-steam-gauge-ars-reveals-steams-most-popular-games/

[30] J. Grubb, J. "Dota 2 grows larger than World of Warcraft, but League of Legends still crushes both," VentureBeat, 2014. [Online]. Available: http://venturebeat.com/2014/05/05/dota-2-grows-larger-than-world-of-warcraft-but-league-of-legends-still-crushes-both/

[31] Bruno. "DotA 2 Replay Parser," 2013. [Online]. Available: http://www.cyborgmatt.com/2013/01/dota-2-replay-parser-bruno/

[32] T. Mahlman. "Dotalys 2," 2014. [Online]. Available: http://www.lighti.de/dotalys-2/

[33] Valve Corporation. "The Source Engine Documentation," 2014. [Online]. Available: https://developer.valvesoftware.com/wiki/SDK_Docs

[34] D. Kotsakos, G. Trajcevski, D. Gunopulos, and C. Aggarwal. "Time series data clustering," in *Data Clustering: Algorithms and Applications*, C. Aggarwal and C. Reddy, Eds. Chapman and Hall/CRC, 2013.

[35] Brandmaier, A. M. *Permutation Distribution Clustering and Structural Equation Model Trees*. Dissertation. Saarland University, 2012.

[36] A. M. Brandmaier. *pdc: Permutation Distribution Clustering*. R package version 0.5, 2014. [Online]. Available: http://CRAN.R-project.org/package=pdc

[37] M. Maechler, P. Rousseeuw, A. Struyf, M. Hubert, and K. Hornik. *Cluster: Cluster Analysis Basics and Extensions*. R package version 1.15.2, 2014.

[38] P. J. Rousseeuw. "Silhouettes: a Graphical Aid to the Interpretation and Validation of Cluster Analysis". *Computational and Applied Mathematics* 20, pp. 53–65, 1987.